\documentclass{article}

\usepackage[preprint]{neurips_2022}

\usepackage[utf8]{inputenc} 
\usepackage[T1]{fontenc}    
\usepackage{hyperref}       
\usepackage{url}            
\usepackage{booktabs}       
\usepackage{amsfonts}       
\usepackage{nicefrac}       
\usepackage{microtype}      
\usepackage{xcolor}         
\usepackage{multirow}
\usepackage{amsmath}
\usepackage{graphicx}
\usepackage{threeparttable}
\usepackage{enumitem}
\usepackage{amssymb}
\usepackage{makecell}
\usepackage{wrapfig}

\title{Spike Calibration: Fast and Accurate Conversion of Spiking Neural Network for   Object Detection and Segmentation}

\author{%
	{Yang Li$^{1,2,3}$\thanks{These authors contributed equally.}\hspace{4pt},  Xiang He$^{1,2,3*}$,  Yiting Dong$^{1,2,4}$,
	Qingqun Kong$^{1,2,3}$, Yi Zeng$^{1,2,3,4,5}$\thanks{Corresponding Author.}}\\
	$^{1}$Institute of Automation, Chinese Academy of Sciences (CAS)\\
	$^{2}$Research Center for Brain-inspired Intelligence, Institute of Automation, CAS \\
	$^{3}$School of Artificial Intelligence, University of Chinese Academy of Sciences\\
	$^{4}$3School of Future Technology, University of Chinese Academy of Sciences\\
	$^{5}$National Laboratory of Pattern Recognition, Institute of Automation, CAS \\

	\texttt{\{liyang2019,hexiang2021,dongyiting2020,qingqun.kong,yi.zeng\}@ia.ac.cn}

}

\begin{document}

	\maketitle

	\begin{abstract}
		Spiking neural network (SNN) has been attached to great importance due to the properties of high biological plausibility and low energy consumption on neuromorphic hardware. As an efficient method to obtain deep SNN, the conversion method has exhibited high performance on various large-scale datasets. However, it typically suffers from severe performance degradation and high time delays. In particular, most of the previous work focuses on simple classification tasks while ignoring the precise approximation to ANN output. In this paper, we first theoretically analyze the conversion errors and derive the harmful effects of time-varying extremes on synaptic currents. We propose the Spike Calibration (SpiCalib) to eliminate the damage of discrete spikes to the output distribution and modify the LIPooling to allow conversion of the arbitrary MaxPooling layer losslessly. Moreover, Bayesian optimization for optimal normalization parameters is proposed to avoid empirical settings. The experimental results demonstrate the state-of-the-art performance on classification, object detection, and segmentation tasks. To the best of our knowledge, this is the first time to obtain SNN comparable to ANN on these tasks simultaneously. Moreover, we only need 1/50  inference time of the previous work on the detection task and can achieve the same performance under  0.492$\times$ energy consumption of ANN on the segmentation task. 
	\end{abstract}

	\section{Introduction}
	\label{intr}
	Although traditional deep learning has achieved superior performance in many fields \cite{silver2016mastering,ye2020mastering,brown2020language}, it still relies on tendons computing resources and energy consumption \cite{silva2021performance,yigitcanlar2021green}, making it difficult to deploy on edge devices. As the third generation artificial neural network (ANN) \cite{maass1997networks,taherkhani2020review}, the spiking neural network (SNN) has high biological plausibility \cite{zhao2021spiking,zhang2018brain} and energy efficiency \cite{roy2019towards}. The spiking neurons receive synaptic current to update the membrane potential and send discrete spikes when it exceeds the threshold. The event-based computation \cite{li2021n} makes the spiking neurons participate in calculations only when they elicit spikes \cite{wunderlich2021event}. Therefore, it exhibits the attributes of low energy consumption and fast inference \cite{park2020t2fsnn}. The excellent performance of spiking neural networks in many kinds of neuromorphic hardware \cite{akopyan2015truenorth,davies2018loihi,pei2019towards} has shown that it is constructive to achieve high performance and low energy consumption artificial intelligence.
	
	However, the unavailability of effective training methods significantly delays the success of SNNs. Due to the non-differentiable characteristic of spikes \cite{zhang2018plasticity}, the traditional backpropagation (BP) algorithm cannot be directly applied to SNN \cite{pfeiffer2018deep}. In recent years, many works based on surrogate gradient have gradually achieved performance comparable to ANN on large-scale datasets\cite{wu2018spatio,shrestha2018slayer,jin2018hybrid,zhao2021backeisnn,fang2021deep}. However, it is still limited to shallow SNNs and simple tasks due to the instability of deep SNNs. In addition,  a lot of storage and computing power is required. The conversion method \cite{cao2015spiking,diehl2015fast} maps the well-trained ANN weights to SNN and achieves high performance while maintaining low energy consumption. However, it typically suffers from severe performance degradation and time delays. Although many methods have been proposed to improve conversion accuracy, such as p-norm\cite{rueckauer2017conversion}, soft reset \cite{han2020rmp}, quantization activation function \cite{yan2021near}, threshold shift\cite{deng2021optimal}, etc., it is still challenging to achieve lossless conversion. In particular, they mainly focus on classification tasks that require only a valid maximum of the output while ignoring more accurate approximations. Spiking-YOLO\cite{kim2020spiking} is the first to explore the detection task, but it requires thousands of time steps to work, and the performance degrades significantly. Furthermore,  SNNs for semantic segmentation still have a big gap with ANNs \cite{kim2021beyond} and can only achieve good results in some small-scale datasets \cite{kirkland2020spikeseg,patel2021spiking}, which can not be applied to the natural scene.
	
\begin{figure}[t]
		\centering
		\includegraphics[scale=0.27]{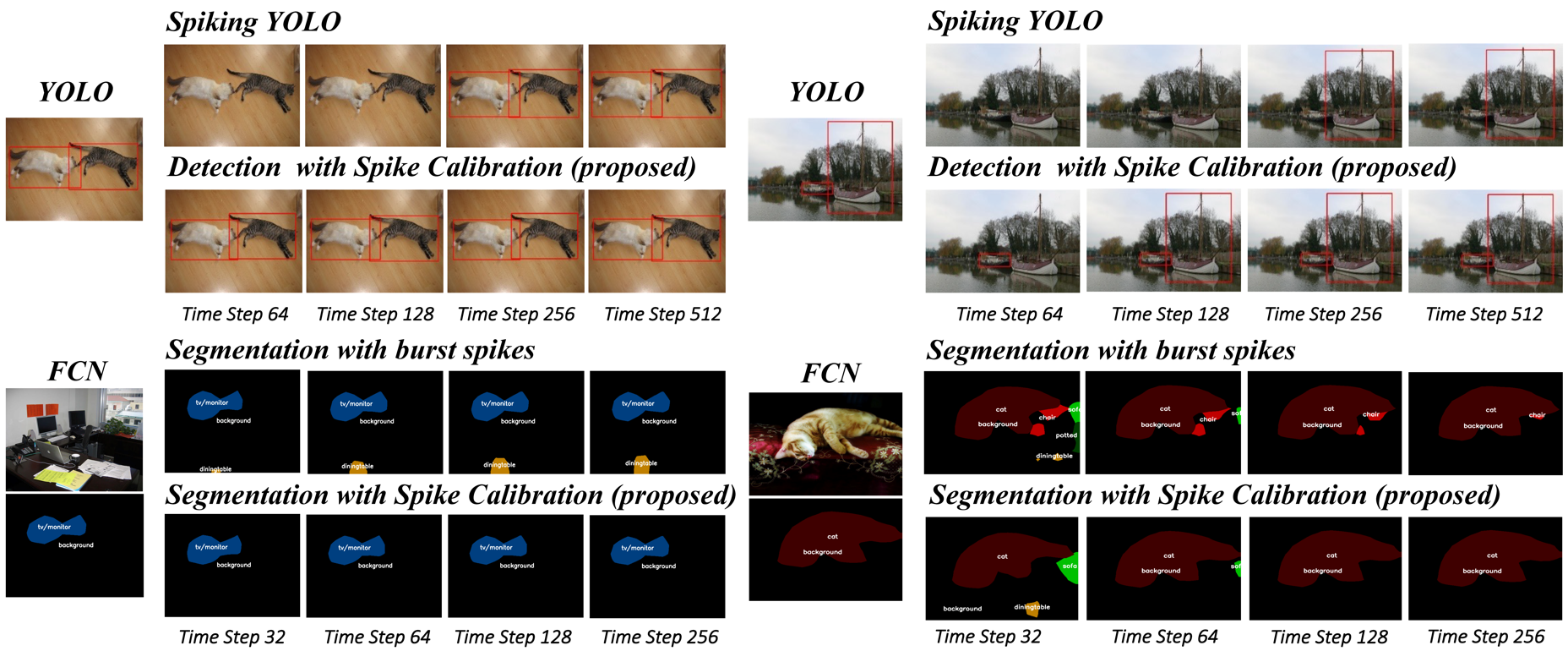}
		\caption{Object detection and segmentation results. (Spiking-YOLO vs. Burst Spikes vs. SpiCalib)}
		\label{intro}
	\end{figure}

	There is still a lack of accurate conversion theory to make converted SNNs capable of object detection or segmentation tasks. In order to make the output distribution of SNN and ANN be the same and then make the converted SNN get the same performance, we dive into analyzing the error of the conversion process and propose the solutions. Fig. \ref{intro} demonstrates that our SNN can achieve almost the same performance as ANN in object detection and segmentation tasks within 128 time steps. Our main contributions can be summarized as follows:
	\begin{itemize}[leftmargin=*]
		\item We theoretically analyze the error of the conversion process and devive them into clipping error, Spikes of Inactivated Neurons (SIN) error, and MaxPooling error. We find that the firing rate is relative to the time-varying maximum of the sum of received synaptic currents rather than the last value, which has been neglected by previous works and can cause some neurons in SNN to fire when they should not be activated.
		\item  We propose spike calibration (SpiCalib) to detect SIN error and correct false spikes by monitoring neurons' average interspike interval (ISI). We also modify LIPooling so that the MaxPooling layer can be converted losslessly. Thus we offer a promising complete conversion framework to achieve high-performance converted SNN.
		\item We evaluate our method for classification, object detection, and segmentation tasks. The proposed method can achieve state-of-the-art performance in a shorter simulation time than other conversion methods. For the first time, we achieve high-performance SNN comparable to ANN on object detection and segmentation tasks on the PASCAL VOC dataset within 128 time steps.
	\end{itemize}
	
	\section{Related Work}
	\label{related}
	Many methods have been proposed to reduce time delays and performance decrease. Diehl et al. \cite{diehl2015fast} propose the data-based and model-based normalization, which enables the firing rate of IF neurons to approach the maximum activation value of each layer of ANN. Rueckauer et al.\cite{rueckauer2017conversion} propose the p-Norm to overcome the problem of large time delay caused by outliers. The soft-reset method is proposed by Han et al. \cite{han2020rmp}, which significantly reduces the information loss of each layer. However, the SNN still has a long time delay. Kim et al. \cite{kim2018deep,park2019fast} propose the concept of weighted spikes, using phase coding to pack more information into the spike to reduce the simulation time. On this basis, bistable spiking neural network \cite{li2021bsnn} improves the accuracy of information transmission by accumulating spikes. In recent work, Deng et al. \cite{deng2021optimal} and Li et al. \cite{li2021free} achieve good performance by migrating the bias. Li et al. \cite{li2022efficient} propose LIPooling to solve the conversion error caused by the MaxPooling layer. Yan et al. \cite{yan2021near} use quantization to make ANN more robust for conversion. However, the above work is limited to simple classification tasks.
	
	Spiking-YOLO \cite{kim2020spiking} is the first to explore detection tasks through channel-Norm and IBT, which still required more than 8000 time steps to work. In the follow-up work\cite{kim2020towards}, the performance has improved by adding two-phase threshold voltages, but it still needs more than 5000 steps. Other works try to hybrid ANN and SNN for detection tasks, but the effect was still not ideal. As for semantic segmentation, Kirkland et al. \cite{kirkland2020spikeseg} use STDP for the image segmentation task, but it can only work in simple datasets. Kim et al. \cite{kim2021beyond} try using the surrogate gradient algorithm on the VOC dataset, but the effect was poor. To further narrow the gap between the ANN and SNN, we thoroughly analyze the reasons for the performance degradation during the conversion and achieve excellent performance on the detection and segmentation tasks.
	\section{Method}
	
	In this section, we first introduce the neuron models of ANN and SNN, then derive the mathematical framework of conversion and go deeper into the error. We propose SpiCalib, MLIPooling, and Bayesian optimization for p for an accurate and fast conversion. Finally, we analyze how they can achieve the optimal result of conversion.
	
	\subsection{ANN-SNN Conversion}
	\label{asc}
	
	\paragraph{Neuron Model for ANN.} For ANN, the forward propagation process can be described as a linear transformation process and a nonlinear mapping of the ReLU activation function:
	\begin{align}
		\textbf{o}^{(\ell)} = g(\textbf{a}^{(\ell)}) = g(\textbf{W}^{(\ell)}\textbf{o}^{(\ell-1)}), \quad 1\leq \ell \leq L,
	\end{align} where $L$ is the total number of layers of the network, $\textbf{o}^{(\ell)}=(o_1^{(\ell)}, o_2^{(\ell)}, \dots, o_{M_{\ell}}^{(\ell)})^T$ and $a^{(\ell)}=(a_1^{(\ell)}, a_2^{(\ell)}, \dots, a_{M_{\ell}}^{(\ell)})^T$ refer to the output and the input vectors of the ReLU funciton $g(x)$ in $\ell$-th layer, $M_{\ell}$ is the number of neurons in $\ell$-th layer. $\textbf{W}^{(\ell)}$ denotes the weight matrix between neurons in layer $\ell-1$ and layer $\ell$. In order to simplify subsequent analysis, we omit the bias $\textbf{b}^{(\ell)}$.
	
	\paragraph{Neuron Model for SNN.} For SNN, we use the IF neuron model. Neurons in layer $\ell$ continuously receive spike input from neurons in layer $\ell-1$ and dynamically update their membrane potential:
	\begin{align}
		{V}_i^{(\ell)}(t) = {V}_i^{(\ell)}(t-1) + \textbf{W}_i^{(\ell)}\textbf{s}^{(\ell-1)}(t),
	\end{align}where $V_i^{(\ell)}(t)$ means the membrane potential of the $i$-th neuron in $\ell$-th layer in time step $t$. $\textbf{s}^{(\ell-1)}=(s_1^{(\ell)}, s_2^{(\ell)}, \dots, s_{M_{\ell}}^{(\ell)})^T$ denotes the spikes vector of the neurons in $\ell-1$. When the membrane potential ${V}_i^{(\ell)}(t)$ exceeds the threshold value ${\theta}_i^{(\ell)}$, the neuron will elicit spikes as follows:
	\begin{align}
		{s}_i^{(\ell)}(t)  = &
		\begin{cases}
			1 \quad if \quad {V}_i^{(\ell)}(t) \geq {\theta}_i^{(\ell)},\\
			0 \quad else.
			\label{spike}
		\end{cases},
		\quad {V}_i^{(\ell)}(t) = {V}_i^{(\ell)}(t) - {\theta}_i^{(\ell)}
	\end{align}
	We further use the burst neuron model in \cite{li2022efficient} to elicit multiple spikes within two time steps, which can greatly improve the information transmission efficiency.  At the same time, to reduce the information loss, we use the soft reset \cite{han2020rmp}. When the neuron spikes, its membrane potential will be subtracted by the threshold ${\theta}_i^{(\ell)}$.
	
	\paragraph{ANN-SNN Conversion.} For ANN-SNN conversion, the key is to use the firing rate of neurons in SNN to approximate the activation value in ANN. If the total simulation time is $T$ and ${r}_i^{(\ell)}(T)=\frac{\sum_{t=0}^{T} {s}_i^{(\ell)}(t)}{T}$ is used to denote the firing rate of the neuron $i$, then we will get
	\begin{align}
		\label{eq4}
		{r}_i^{(\ell)}(T) = \frac{\sum_{t=0}^{T} \textbf{W}_i^{(\ell)}\textbf{s}^{(\ell-1)}(t)}{T{\theta}_i^{(\ell)}} - \frac{{V}_i^{(\ell)}(T)}{T} + \frac{{V}_i^{(\ell)}(0)}{T}.
	\end{align}
	Usually, the threshold ${\theta}_i^{(\ell)}$ is set to $1$. When the simulation time $T$ is long enough, $\frac{{V}_i^{(\ell)}(T)}{T} \approx 0$ and $\frac{{V}_i^{(\ell)}(0)}{T} \approx 0$. Then ${r}_i^{(\ell)}(T) \approx {o}_i^{(\ell)}$ under the condition of ${r}_i^{(\ell-1)}(T) = {o}_i^{(\ell-1)}$. However, the firing rate is limited in $[0,1]$. Thus we follow the weight normalization method \cite{rueckauer2017conversion} to scale the activation values. Specifically, we use the $p$-th percentile of the activation values in the $\ell$-th layer $\max^{(\ell)}_p=\operatorname{sort}(\textbf{o}^{(\ell)})[\operatorname{len}(\textbf{o}^{(\ell)})*p]$ to scale the weight, where $p$ is a parameter as follows:
	\begin{align}
		\hat{\textbf{w}}^{(\ell)} = \textbf{w}^{(\ell)}
		\frac{\max^{(\ell-1)}_p}{\max^{(\ell)}_p},\quad \hat{\textbf{b}}^{(\ell)} = \frac{\textbf{b}^{(\ell)}}{\max^{(\ell)}_p}.
	\end{align}
	
	\subsection{Conversion Error Analysis}
	\label{ce}
	\begin{figure}[t]
		\centering
		\includegraphics[scale=0.25]{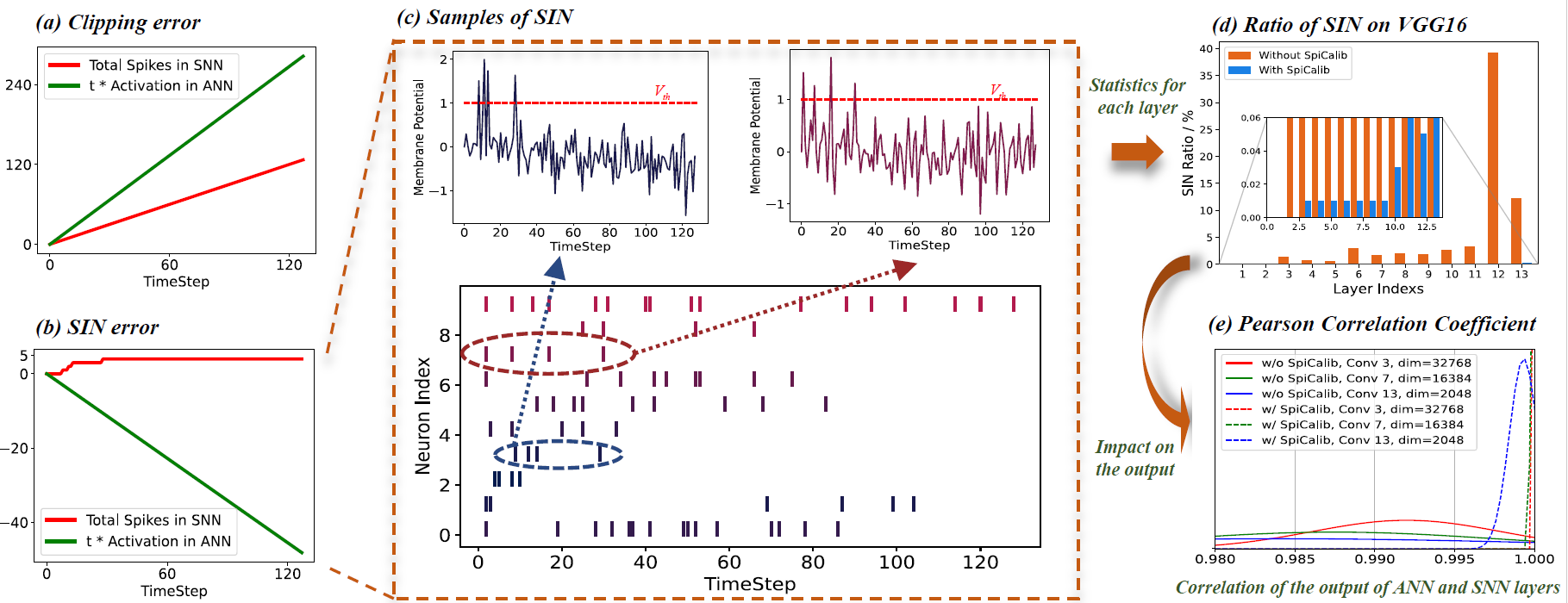}
		\caption{The conversion errors and their influence on the distribution of SNN. (a) The total synaptic current received by the neuron will be more and more different from the product of the activation value (>1) and time. (b) When $\sum _{\tau=0}^t {I}_i^{(\ell)}(\tau) > \theta_i^{(\ell)}$, the neuron will spike, regardless of whether the corresponding activation value is positive or negative. (c) SIN problem usually presents multiple spikes at the beginning of simulation and silence at the later stage. (d) As the network layer deepens, SIN errors occur in more and more neurons that should not be activated. (e) Due to the errors in the values of multiple nodes, the distribution of the hidden layer of SNN is quite different from that of ANN, which directly affects the performance of SNN in tasks.}
		\label{err}
	\end{figure}

	Although weight normalization is used to improve inference speed, some neurons still have an activation value greater than 1. Even with burst spikes, neurons with activation values above the maximum spikes are not effectively approached by IF neurons, as shown in Fig. \ref{err} (a).
	Therefore, the index of neurons at layer $\ell -1$ can be divided into $\boldsymbol{R}_1= \left\{j \mid \frac{\textbf{o}_{j}^{(\ell-1)}}{max_{p}^{(\ell-1)}} \leq \Gamma \right\}$ and $\boldsymbol{R}_2= \left\{j \mid \frac{\textbf{o}_{j}^{(\ell-1)}}{max_{p}^{(\ell-1)}} \geq \Gamma \right\}$, where $\Gamma$ is the number of burst spikes. Then, the total synaptic current received by neuron $i$ in $\ell$-th layer is
	\begin{align}
		\label{dilema}
		\sum_{\tau}^{t} {I}_{i}^{(\ell)}(\tau)
		=\sum_{\tau}^{t} \sum \hat{{W}}_{i j}^{(\ell)} {s}_{j}^{(\ell-1)}(\tau)
		=\sum_{j \in \boldsymbol{R}_1} \frac{{W}_{i j}^{(\ell)}}{max_{p}^{(\ell)}}  \sum_{\tau}^{t} {s}_{j}^{(\ell-1)}(\tau)+\sum_{j \in \boldsymbol{R}_2} \frac{{W}_{i j}^{(\ell)}}{max_{p}^{(\ell)}}  t,
	\end{align} where $\tau$ represents the time step between $0$ and $t$. Eq. \ref{dilema} shows the dilemma of speed and accuracy in conversion. $p$ is usually an empirical value between 0.9 and 0.9999. The process brings some errors to the conversion.
	
	The current conversion works use real-value input and use the first layer as the encoding layer. This causes the synaptic current $I$ received by  IF neurons in the first layer to be a constant value and $\forall t \in [0, T], {I}_i^{(1)}(t)=\textbf{W}_i^{(1)}\textbf{o}^{(1)}$. According to Eq. \ref{eq4}, if $\boldsymbol{R}_2 \in \varnothing$, the firing rate of the first layer can ideally equal the activation value in ANN with enough long time steps, and get
	\begin{align}
		{r}_i^{(1)}(T)=\left\lfloor\frac{\sum_{t=0}^T {I}^{(1)}_i(t)}{T}\right\rfloor
	\end{align}
	As neurons receive discrete spikes after the encoding layer, the synaptic current becomes extremely unstable, and the interspike interval (ISI) is distributed in various shapes, but its mean value should be $\frac{1}{\textbf{o}_i^{(\ell)}}$. We use floor function $\lfloor{x\rfloor}$ to return the greatest integer that less than or equal to $x$, $\kappa_t = \sum _{\tau=0}^t {I}_i^{(\ell)}(\tau)$ represent the sum of synaptic currents received by the neuron $i$ in $\ell$-th layer by time t. Let us set the sequence $K=[\lfloor{\kappa_0 \rfloor}, \lfloor{\kappa_1 \rfloor}, \dots, \lfloor{\kappa_T \rfloor}]$, then we find that\textbf{\textit{ the total number of spikes ${S}_i^{(\ell)}$ emitted is the time-varying extremum of K , rather than the last element of K.}} Neglect the floor error,  K$_T=\lfloor{\kappa_T \rfloor}=T o_i^{(\ell)}$, which can be discribed as 
	\begin{align}
		{S}_i^{(\ell)}(T) = \operatorname{ext}\{ K_t | 0\leq t\geq T\} = \max \limits_{0\leq t\leq T}\{\lfloor{\sum\limits _{\tau=0}^t {I}_i^{(\ell)}(\tau) \rfloor}\}
	\end{align} where $\operatorname{ext}\{K_t\}$ denotes the maximum of the temporal sequential. For neuron $i$ in SNN, if ${o}_i^{(\ell)}\geq 0$, the fluctuation of synaptic current will only make the variance of ISI distribution bigger, but will not affect its mean value. However, if ${o}_i^{(\ell)}< 0$ and synaptic currents exceeds the threshold in a local time, neurons that should not be activated in the corresponding ANN could still emit spikes, call spikes of the inactivated neuron (SIN). And the effect cannot be eliminated through longer simulation, as is shown in Fig. \ref{err} (b) and Fig. \ref{err} (c). Note that the number of SIN is uncertain. Suppose each neuron in $\boldsymbol{R}_3 = \{j|{S}_j^{(\ell-1)}>0, {o}_j^{(\ell-1)}<0\}$ emits one spike, the $T$ is long enough, the floor error could be neglected and there is no clipping error, the error $E_i^{(\ell)}={r}_i^{(\ell)}(T) - {o}_i^{(\ell)} $ of neuron $i$ in layer $\ell$ ($o_i^{(\ell)}>0$) will be
	\begin{align}
		E_i^{(\ell)} = \frac{{S}_i^{(\ell)}(T)}{T} - o_i^{(\ell)} =  \frac{1}{T}\left(\max \limits_{0\leq t\leq T}\{\lfloor{\sum\limits _{\tau=0}^t {I}_i^{(\ell)}(\tau) \rfloor}\}-K_T \right) \approx \frac{1}{T} \lfloor\sum_{j\in \boldsymbol{R}_3} {W}_{ij}^{(\ell)}\rfloor
	\end{align} 
	
	Fig. \ref{err} (d) demonstrate the ratio of SIN on VGG16 on CIFAR100. As the layer goes deeper, the SIN problem becomes more and more serious and nonnegligible. The distribution of firing rate in SNN and the activation value in ANN offers high variance and relatively high mean value, as shown in Fig. \ref{err} (e). A high mean value indicates that the magnitude relationship of $\textbf{r}^{(\ell)}$ is consistent with ANN, while a high variance indicates that each value in $\textbf{r}^{(\ell)}$ has a high probability of deviating from the corresponding activation value. It is also why the current conversion method cannot be effectively applied to object detection and segmentation tasks.
	
	As described in \cite{li2022efficient}, the output of the MaxPooling layer of converted SNN is usually greater than the actual maximum value. So previous works use Average pooling instead. Although LIPooling effectively solves the above problems, it seems to work only when the kernel size equals the stride. Moreover, lossless conversion is still impossible if the pooling layer uses ceil mode.

	\subsection{Spike Calibration}
\begin{figure}[!t]
		\centering
		\includegraphics[scale=0.26]{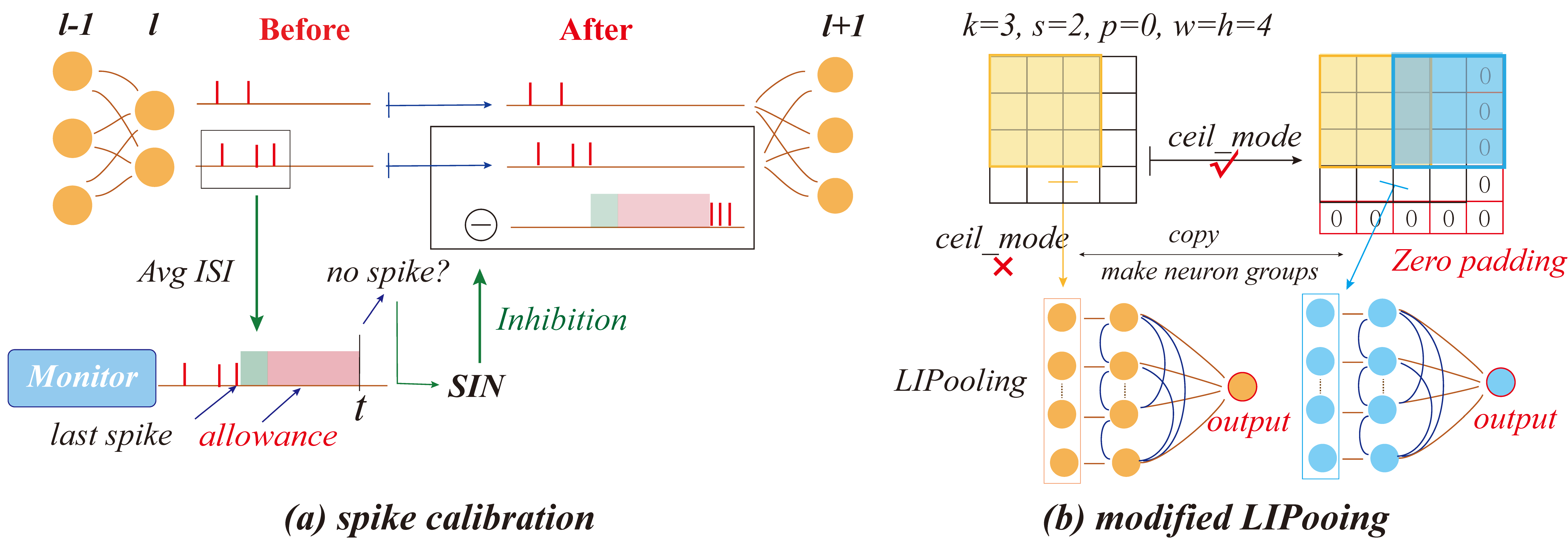}
		\caption{Our proposed methods. (a) We monitor the average ISI of the spiking neurons online and correct the influence of the false spikes on the output distribution by suppressing the historical output of the neurons that do not spike after a certain threshold; (b)  An independent lateral inhibition system for each pooling block is established by duplicating the neurons' spikes to prevent the neurons from being affected by the action of multiple pooling blocks.}
		\label{method}
	\end{figure}

	By using the burst mechanism in \cite{li2022efficient}, the clipping error can be greatly reduced.
	During SNN's inference, it is impossible to know which neuron in the corresponding ANN should be inactivated. Interspike interval (ISI), as an essential attribute of SNN, can be used to evaluate the activity characteristics of neurons. Ideally, ISI corresponds strictly to the activation value ${o}_i^{(\ell)}$, for example, the first layer of SNN, but in other layers, its value fluctuates around the mean due to the influence of discrete spikes. We find that neurons with SIN usually fire early and then silence during the conversion process. Based on the discovery, as shown in Fig.\ref{method} (a), we make a monitor to calculate the average ISI of neurons $\boldsymbol{\phi}_i^{(\ell)}(t)=(\phi_1^{(\ell)}, \phi_2^{(\ell)}, \dots, \phi_{M_{\ell}}^{(\ell)})^T$ online:
	\begin{align}
		\phi_i^{(\ell)}(t) = 
		\begin{cases} 
			\frac{\phi_i^{(\ell)}(t) * N + t - t^{'}}{N+1}, \quad \operatorname{if} \quad s_i^{(\ell)}(t)\neq 0,\\ 
			\phi_i^{(\ell)}(t) , \quad else,
		\end{cases}
	\end{align} where $N$ denotes the number of spikes, $t^{'}$ means the time of last spike, with the initial value of 0. The neuron is expected to elicit the next spike within the average ISI. When the monitor finds the spike is still not released after the allowance $\beta$, we think there happens a SIN error. Another synapse connected by the neuron is involved in the transmission of information. It connects neurons $i$ to neurons $j$ like synapse $W_{ij}^{(\ell)}$, but has a  weight of  $-W_{ij}^{(\ell)}$. The twin synapse will continue to deliver spikes that counteract the fault spikes until its cumulative effect on the neuron in the next layer disappears. This mechanism ensures the detection and calibration of SIN problems. Then, the firing rate of neuron $i$ can be expressed as the follows and the converison error will be 0 according to Eq. \ref{eq4}. Here $s_{neg}^{(\ell)}$ denotes the spikes deliveried by the twin synapse.
	\begin{align}
		r_i^{(\ell)}(T) = \frac{\max \limits_{0\leq t\leq T}\{\lfloor{\sum\limits _{\tau=0}^t {I}_i^{(\ell)}(\tau) \rfloor}}{T} - \frac{\sum \limits _{t=0}^T \textbf{W}_i^{(\ell)}\textbf{s}_{neg}^{(\ell)}(t)}{T} \approx \frac{K_T}{T}= o_i^{(\ell)}
		\end{align}
	
	\subsection{MLIPooling}
	The core of LIPooling is to provide a mutual suppression process for neurons in pooling blocks and finally output the sum of the activities of all neurons. We define $k$ the kernel size and $s$ the stride. If ceil mode is used to sample, the number of neurons in the pooling block is less than $k*k$. To make LIPooling work in the scene, as shown in Fig.\ref{method} (b), we fix the right side and the bottom of the original feature map with zero padding. In addition, when $k\neq s$, the same neuron will participate in the inhibition of multiple pooling blocks simultaneously, resulting in the pooling layer not working correctly. We first copy a separate region for each pooling block to ensure that the neurons involved in calculating multiple pooling blocks can participate in the calculation of their respective regions to solve the interference. According to \cite{li2022efficient}, the MaxPooling layer can be converted with MLIPooling losslessly.
	
	\subsection{Bayesian Optimization}
	Even with the use of the Burst mechanism, the weight normalization parameter $p$ is still important. We use KL divergence to represent the information loss caused by the output vector of SNN when it represents the real distribution of ANN's output, which can be written as 
	\begin{align}
		KL(\textbf{r}(\mathcal{D}|p) || \textbf{o}(\mathcal{D} | p)) = \sum_i {r}_i(\mathcal{D}|p) \log{\frac{{r}_i(\mathcal{D}|p)}{{o}_i(\mathcal{D}|p)}},
	\end{align} where $\mathcal{D}$ denotes the training datasets, $p$ is the p-norm parameter. We hope that the KL divergence between the two distributions will be used as the evaluation index to select the optimal parameter $p$,
	\begin{align}
		\arg \min \limits_p  KL(\textbf{r}(\mathcal{D}|p) || \textbf{o}(\mathcal{D} | p))
	\end{align}
	Then, we use the Bayesian Optimization Algorithm (BOA) to solve the above optimization problem. The Gaussian process and expected improvement (EI) are used for the prior and acquisition functions, respectively, which are both effective methods for finding the optimal value with little evaluation. We randomly select a batch for evaluation because the training data satisfy the dependently identically distribution (i.i.d). The method can effectively avoid empirical selection parameters.

	\section{Experiments and Results}
	\label{exp}
	In this section, all experiments are performed on NVIDIA A100 and based on Pytorch platform. For classification task, we report the results of the proposed SpiCalib on CIFAR10, CIFAR100 \cite{krizhevsky2009learning} and ImageNet \cite{russakovsky2015imagenet}. Then we test the YOLO \cite{redmon2016you} based on VGG16 and ResNet50 backbones on PASCAL VOC \cite{everingham2010pascal}, and COCO datasets \cite{lin2014microsoft} and compare it with Spiking-YOLO and its improvement. We also convert the FCN \cite{long2015fully} network for semantic segmentation. In ANN, data augmentation and batch normalization are used. In SNN, direct input strategy is used and the batch size is set to 50 for Bayesian optimization. For burst spike, we follow \cite{li2022efficient} and use $\Gamma=5$. See Supplementary Material for detailed experimental configuration, overall proposed algorithm and the result of classification. 
	\begin{figure}[t]
		\centering
		\includegraphics[scale=0.75]{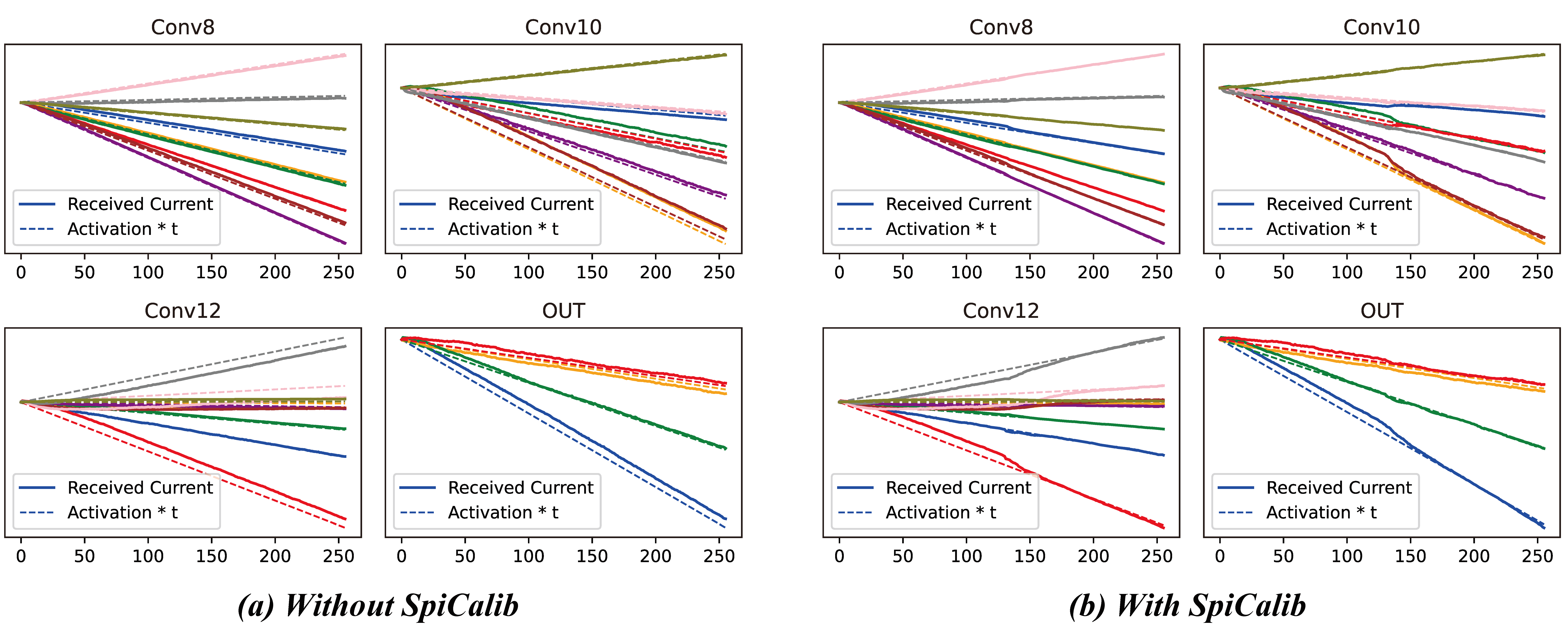}
		\caption{Visualization of the total synaptic current received by neurons at each layer. Each color represents a different neuron, and the ordinate is the sum of synaptic currents received up to time step t. SpiCalib could eliminate the effect of false spikes on information transmission. }
		\label{hihe}
	\end{figure}
	
	\begin{table}[!t]
		\centering
		\caption{Experimental results on CIFAR100 on VGG16}
		\label{ablation}	
		\begin{threeparttable}
			\resizebox{\linewidth}{!}{
				\begin{tabular}{c|c|ccccc|c}
					\toprule
					Method &ANN &SNN Best & T=32 & T=64 & T=128 & T=256 & KLDiv (T=256)\\
					\midrule
					p-Norm \cite{rueckauer2017conversion} & 78.49& 58.44 & 44.88 & 51.89 & 56.02 & 58.44 & 15277.852539 \\
					Channel-Norm\cite{kim2020spiking} & 78.49 & 74.74 & 54.03 & 67.34 & 72.50 & 74.73 &\multirow{6}{*}{NA} \\
					Spike-Norm\cite{sengupta2019going} & 71.22 & 70.77& - & - & - & - & \\
					TSC\cite{han2020deep}  &71.22 & 70.97 &- &- & 69.86 & 70.65 &\\
					RMP-SNN\cite{han2020rmp} &71.22 & 70.93 &- &- &63.76 & 68.34&\\
					Opt.\cite{deng2021optimal} & 77.89& 77.71 & 7.64 & 21.84 & 55.04 & 73.54&\\
					Calibration\cite{li2021free} & 77.89 & 77.87 & 73.55 & 76.64 & 77.40 & 77.68&\\
					\midrule
					Burst \cite{li2022efficient} (p=0.95) & 78.49 & 74.67 & 73.72 & 74.40 & 74.57 & 74.62 & 4780.952148\\
					Burst \cite{li2022efficient} (p=0.999)  & 78.49& 77.70 & 62.93 & 74.31 & 76.77 & 77.68 & 573.957031 \\
					Burst\tnote{*} \cite{li2022efficient} (BOA, p=0.995)  & 78.49& 78.12 & 71.54 & 76.57 & 77.71 & 78.09 & 293.472595 \\
					{ Burst + MLIPooling} & 78.49 & {78.63} & {74.00} & {77.94} & {78.59} & {78.59} & {53.483173} \\
					Burst + SpiCalib (64) &78.49  & 77.97 & 71.54 & 76.57 & 77.63 & 77.93 & 415.863342 \\
					Burst + MLIPooling + SpiCalib (64) &78.49 & 78.65 & 74.00 & 77.94 & 78.53 & 78.46 & 13.985744 \\
					\textbf{Burst + MLIPooling + SpiCalib (128)} &78.49 & \textbf{78.63} & \textbf{74.00} & \textbf{77.94} & \textbf{78.59} & \textbf{78.48} & \textbf{4.738684} \\
					\bottomrule
			\end{tabular}}
			\begin{tablenotes}   
				\footnotesize           
				\item[*] The subsequent experiments are based on $p$ with BOA.
			\end{tablenotes} 
		\end{threeparttable}
		\label{table1}
	\end{table}
	
	\subsection{Ablation Experiments on Classification Tasks}
	\label{ablationsec}
	We use the proposed algorithm on the CIFAR100 dataset. As mentioned above, accurate conversion requires that the output distribution of each layer of the network approximate exactly to ANN. We visualize the total synaptic current received by IF neurons in the hidden layer and output layer of SNN on CIFAR100 and VGG16 as shown in Fig \ref{hihe}. The dotted lines indicate that the ideal IF neuron receives a steady synaptic current as the first layer does. Over time, the solid and dashed lines will become parallel or at an angle. We think it is because IF neurons receive additional short-term inputs such as SIN or long-term more or less constant stimuli. We set the allowance $\beta = \frac{1}{2}T$. As shown in Fig. \ref{hihe} (b), the total current received by the neuron returns to the same level as the ideal. Then, according to the Eq. \ref{eq4}, we can therefore achieve an accurate conversion.

	We then examine the influence of Bayesian optimization and MLIPooling on the experimental results. It can be seen from Tab. \ref{table1} that Bayesian optimization can find the optimal normalized parameters at a small cost and improve the network's performance. Since LIPooling cannot transform the MaxPooling layer where $k\neq s$, it cannot be compared with LIPooling here. However, MLPooling can rapidly reduce the KL divergence of the output. And then, SpiCalib can further improve the accuracy of the conversion. We find that MLIPooling plays a role in coarse-tuning while SpiCalib plays a role in fine-tuning. That is why the performance with only SpiCalib in the table decreases instead. The comparison with other advanced conversion methods \cite{kim2020spiking,sengupta2019going,han2020deep,han2020rmp,deng2021optimal,li2021free,li2022efficient} also shows the advantages of our method.

		\begin{figure}[!t]
		\centering
		\includegraphics[scale=0.67]{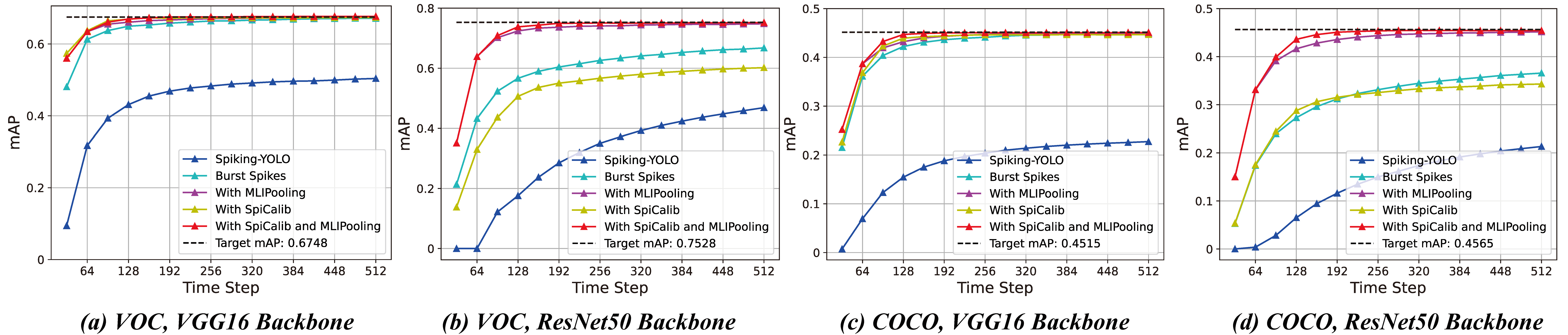}
		\caption{The mAP curves on VOC and COCO datasets based on VGG-16 and ResNet50 backbone.}
		\label{detcurve}
	\end{figure}
	
	\begin{table}[t]
		\centering
		\caption{Comparison of mAP(\%) for PASCAL VOC with various methods.}
		\label{detection}	
		\resizebox{\linewidth}{!}{
			\begin{tabular}{c|ccccc|ccccc}
				\toprule
				\multirow{3}{*}{Methods}  &  \multicolumn{5}{c|}{Backbone-VGG16 }  & \multicolumn{5}{c}{Backbone-ResNet50 }    \\
				\cmidrule{2-11}
				& ANN &64 & 128 & 256 & 512  &ANN&  64 & 128 & 256 & 512\\
				\midrule
				Channel Norm \cite{kim2020spiking} &67.48& 31.74& 43.11 & 48.29 & 50.39  &75.28&  0 & 17.59 & 37.22 & 46.88\\
				Two-phase threshold \cite{kim2018deep} &53.01 &-&-&$<$46.66&-&&&&&\\
				Burst Spikes \cite{li2022efficient}&67.48 & 61.30 & 64.95 & 66.40 & 67.11  & 75.28&  43.23 & 56.65 & 62.61 & 66.76\\
				Burst + MLIPooling &67.48 & 63.43 & 66.06 & 67.04 & 67.42  & 75.28&  63.93 & 72.43 & 74.07 & 74.79\\
				Burst + SpiCalib  &67.48& \textbf{63.69 }& \textbf{66.89} & 67.27 & 67.37  &75.28&   32.99 & 50.63 & 56.67 & 60.17\\
				Burst + MLIPooling + SpiCalib&67.48 & 63.43 & 66.08 & \textbf{67.54 }& \textbf{67.69}  & 75.28&  \textbf{63.85} & \textbf{73.78} & \textbf{75.07} & \textbf{75.21}\\
				\bottomrule
		\end{tabular}}
	\end{table}

	\begin{table}[!t]
		\centering
		\caption{Comparison of mAP(\%) for MS COCO with various methods.}
		\label{detection-coco}	
		\resizebox{\linewidth}{!}{
			\begin{tabular}{c|ccccc|ccccc}
				\toprule
				\multirow{3}{*}{Methods}  &  \multicolumn{5}{c|}{Backbone-VGG16}  & \multicolumn{5}{c}{Backbone-ResNet50 \tnote{2} } \\
				\cmidrule{2-11}
				&ANN &  64 & 128 & 256 & 512  &ANN&  64 & 128 & 256 & 512\\
				\midrule
				Channel Norm \cite{kim2020spiking} &45.15& 6.93& 15.46 & 20.37 & 22.75  &45.65&  0.39 & 6.52 & 14.99 & 21.33\\
				Two-phase threshold \cite{kim2018deep} &26.24 &-&-&$<$21.05&-&&&&&\\
				Burst Spikes \cite{li2022efficient}&45.15 & 36.13 & 42.22 & 44.12 & 44.86  &45.65&  17.39 & 27.30 & 33.13 & 36.58\\
				Burst + MLIPooling  &45.15& 38.68 & 43.21 & 44.70 & 44.94  &45.65&  33.08 & 41.65 & 44.38 & 45.17\\
				Burst + SpiCalib &45.15 & 36.83& 43.94 & 44.62 & 44.66  &45.65&  17.55 & 28.77 & 32.53 & 34.30\\
				Burst + MLIPooling + SpiCalib &45.15& \textbf{38.68} & \textbf{44.70} & \textbf{45.06 }& \textbf{45.10}  &45.65& \textbf{ 33.12} & \textbf{43.61} & \textbf{45.36} & \textbf{45.42}\\
				\bottomrule
		\end{tabular}}
	\end{table}

	\subsection{Object Detection}
	Though some work\cite{chakraborty2021fully, miquel2021retinanet,kim2020spiking} demonstrate the feasibility of designing a fully spiking network for object detection, SNN still suffers from apparent performance degradation. We convert YOLO\cite{redmon2016you} to verify the effectiveness of the proposed conversion method. Fig. \ref{detcurve} shows our experimental results on VOC \cite{everingham2010pascal}and COCO datasets \cite{lin2014microsoft}. Since our network uses ReLU activation functions, only channel-norm is used for testing Spiking-YOLO. When both SpiCalib and MLIPooing are applied, our SNN could achieve 67.69\% and 75.21\% mAP on the VOC dataset with VGG16 and ResNet50 networks and 45.10\% and 45.21\% mAP on the COCO dataset, respectively. Apparently, our performance surpasses that of Spiking-YOLO \cite{kim2020spiking}]and its improvement \cite{kim2020towards} and achieves state-of-the-art performance. The detailed data is shown in the Tab. \ref{detection} and Tab. \ref{detection-coco}. It is worth noting that we achieve the same performance loss in 1/50 time steps of Spiking-YOLO. We find that results with SpiCalib only perform worse than baseline when using ResNet as the backbone. As described in Section \ref{ablationsec}, we think it is because the correction effect of SpiCalib on the already broken distribution is inefficient and amplified by shortcut connection. Fig. \ref{intro} illustrates the efficiency of Spicalib for detection tasks. It demonstrates the advantages of SpiCalib in accurately detecting multiple and small targets in a short time. While Spiking-YOLO needs 512+ time steps to complete effective detection.

	\subsection{Semantic Segmentation}
	
	\begin{table}[t]
		\centering
		\caption{Comparison of results on sementic segmentation.}
		\label{segmen}	
		\resizebox{\linewidth}{!}{
			\begin{tabular}{c|c|cccc||c|cccc}
				\toprule
				Method & Metrics & ANN & T=64 & T=128& T=256 & Metrics & ANN & T=64 & T=128& T=256\\
				\midrule
				\multirow{2}{*}{Burst } & pixel acc.& 89.09 & 87.81 & 88.57 & 88.75& mean IoU & 59.54  & 55.76 & 58.24 & 58.62  \\
				& mean acc.& 74.24  & 70.85 & 72.62 & 72.85  & f.w. IoU & 81.49  & 78.31 & 80.79 & 81.08  \\
				\midrule
				\multirow{2}{*}{Burst + MLIPooling } & pixel acc.& 89.09 & 87.74 & {88.86} & {89.07} & mean IoU & {59.54}  & {55.04} & {58.76} & {59.45} \\
				& mean acc. & {74.24}  & {69.86} & {73.38} & {74.08}& f.w. IoU & {81.49} & {79.16} & {81.01} & {81.40} \\
				\midrule
				\multirow{2}{*}{Burst + Spicalib(128) }& pixel acc.& 89.09 & 87.81 & 88.58 & 88.66 &mean IoU & 59.54  & 55.76 & 28.24 & 58.46  \\
				& mean acc.& 74.24 & 70.85 & 72.62 & 73.24 & f.w. IoU& 81.49  & 75.40 & 80.79 & 81.04  \\
				\midrule
				\multirow{2}{*}{\makecell{\textbf{Burst + MLIPooling }\\\textbf{+ Spicalib(128)}}} & pixel acc.& \textbf{89.09}  & \textbf{87.74} & \textbf{88.86} & \textbf{89.07} &mean IoU & \textbf{59.54}  & \textbf{55.04} & \textbf{58.76} & \textbf{59.56} \\
				& mean acc.& \textbf{74.24}  & \textbf{69.86} & \textbf{73.38} & \textbf{74.66}& f.w. IoU& \textbf{81.49}  & \textbf{79.17} & \textbf{81.01} & \textbf{81.49} \\
				\bottomrule
		\end{tabular}}
	\end{table}
	
	To validate the effectiveness of our proposed conversion method, We also explore the converted SNN for semantic segmentation tasks. Fully Convolutional Network (FCN)\cite{long2015fully}, one of the first deep learning works for semantic image segmentation, is chosen as our conversion target ANN. The pixel accuracy, mean accuracy, mean IoU, and frequency weighted IoU (f. w. IoU) are used as evaluation indexes. Tab. \ref{segmen} shows the improvement of our method over the baseline, with parentheses representing the allowance of Spicalib. It can be seen that MLIPooling can improve the baseline from 72.85\% mean accuracy to 74.08\%, and from 58.62\% mean IoU to 59.45\%. Although the improvement is undeniable, there is still some gap with the target ANN. After using the SpiCalib mechanism, all indicators show that the effect of object segmentation of SNN can be equivalent to that of ANN. Fig. \ref{intro} shows that pixel-wise classification was prone to misclassification with only Burst and could not be compensated by increasing the simulation time. 	 However, SpiCalib can correct the previous misclassification of the pixels and ensure  \begin{wrapfigure}[12]{b}{0.32\textwidth}
		\begin{center}
			\includegraphics[width=0.31\textwidth]{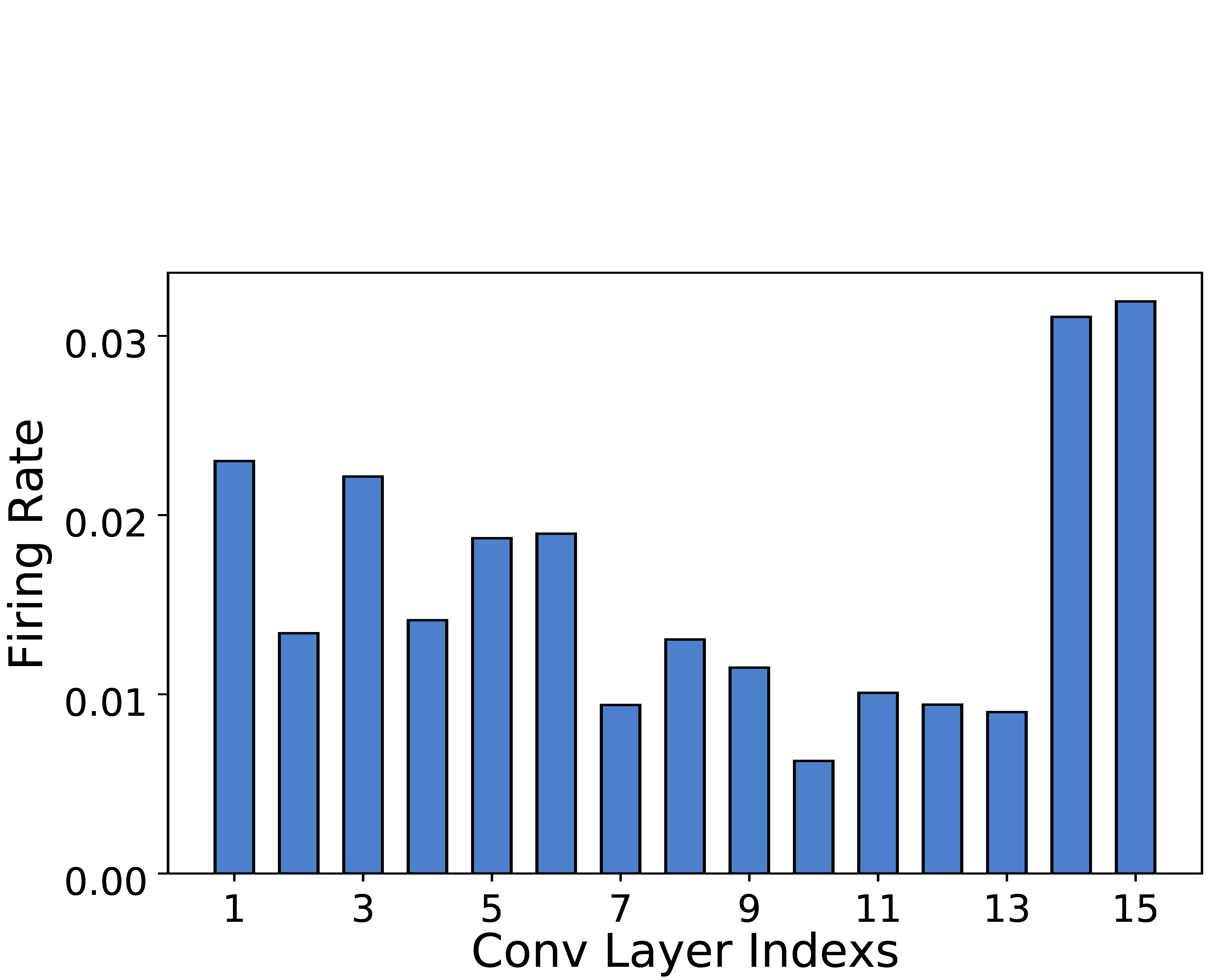}
		\end{center}
		\caption{Firing rate of Spiking FCN on VOC dataset.}
		\label{sparsity}
	\end{wrapfigure}  the accuracy of semantic segmentation.
	
	\subsection{Sparsity and Energy Efficiency}
	In this section, we verify the sparsity of our Spicalib. We choose the FCN network on VOC  with $T=64$. We test the result with all methods we proposed and record the firing rate of SNN on the whole test dataset. As shown in Fig.\ref{sparsity}, the firing rate varies from 0.0062 and 0.0319, respectively, indicating the sparsity of our network. We then quantitatively calculate the energy efficiency of SpiCalib. The spiking neural network shows the energy-efficient property in neuromorphic hardware because neurons only participate in the calculation when they spike. According to the energy calculation formula in \cite{rathi2020diet}, our method consumes only 0.492$\times$ energy of ANN, which shows the efficiency of our method.
	
	\section{Discussion}
	\label{diss}
	In this paper, we present a complete conversion framework for accurate and fast conversion with less power consumption. We theoretically analyze the error of the conversion process and illustrate the influence of SIN on the accurate conversion, which is different from simple classification by finding the maximum. Then, we propose SpiCalib to eliminate the influence of false spikes on SNN's output distribution and modify LIPooling so that any MaxPooling can be converted losslessly in SNN. We also propose using the Bayesian optimization method to select the optimal parameter $p$. We achieve state-of-the-art performance in classification, object detection, and segmentation tasks. Our work can make large-scale SNNs competitive on complex tasks with neuromorphic hardware, which is beneficial for SNN development. Although the improvement on complex tasks is significant, it is still impossible to complete the zero error conversion. The SpiCalib proposed is adequate for most SIN problems, but we find that for a small number of neurons whose activation value is in the minimal immediate domain of zero, they can keep firing throughout the whole simulation time and cannot be corrected by the SpiCalib mechanism. In the future, if we introduce quantization techniques and combine them with SpiCalib, it will make the conversion framework more practice and efficient.

	

\end{document}